\documentclass[letterpaper, 10 pt, conference]{ieeeconf}  

\usepackage{graphicx}
\usepackage[font=small]{caption}
\usepackage{cite}
\usepackage{float}
\usepackage{siunitx}
\usepackage{hyperref}
\usepackage{amsmath}
\usepackage{siunitx}
\usepackage{url}

\hypersetup{colorlinks=true,urlcolor=blue}

\IEEEoverridecommandlockouts                              

\title{\LARGE \bf
NeRF2Real: Sim2real Transfer of Vision-guided Bipedal Motion Skills using Neural Radiance Fields
}

\author{Arunkumar Byravan$^{1}$, Jan Humplik$^{1}$, Leonard Hasenclever$^{1}$, Arthur Brussee$^{1}$, Francesco Nori, \\ Tuomas Haarnoja, Ben Moran, Steven Bohez, Fereshteh Sadeghi, Bojan Vujatovic and Nicolas Heess \\ \small{DeepMind, $^{1}$Equal Contributions} \vspace{-0.7cm}
}

\let\oldtwocolumn\twocolumn
\renewcommand\twocolumn[1][]{%
    \oldtwocolumn[{#1}{
    \begin{center}
            \captionsetup{type=figure}
            \includegraphics[width=\textwidth]{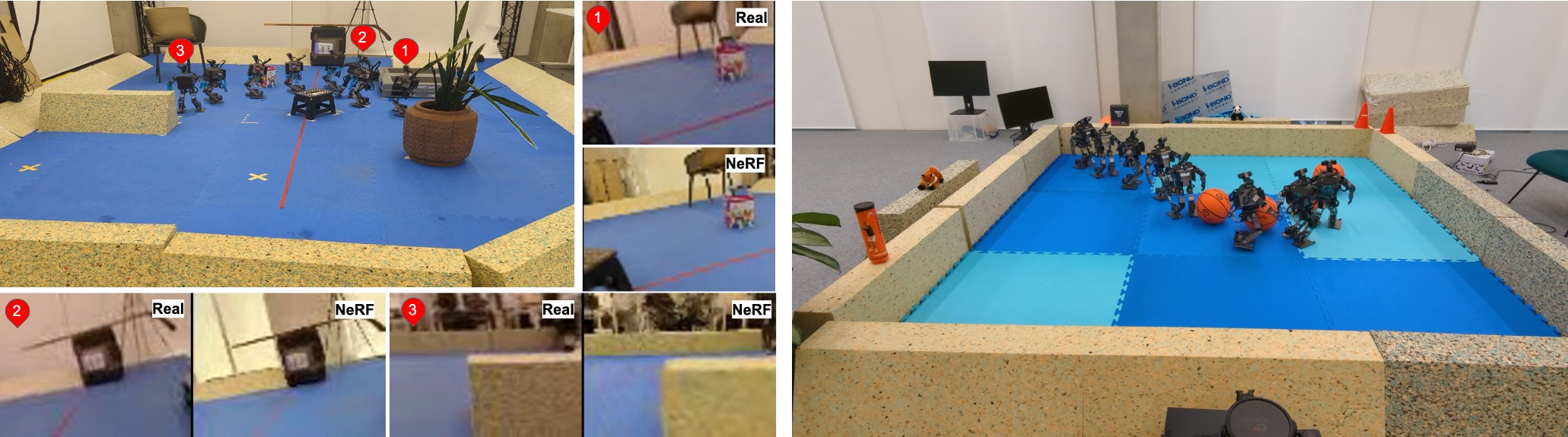}
            \captionof{figure}{
           Zero-shot sim2real transfer results of vision-based bipedal locomotion policies trained using reinforcement learning in two separate simulations created using our NeRF2Real setup. \textbf{Left:} time lapse of a transfer result on a navigation task with a comparison of the robot's head-mounted camera views vs NeRF renderings: `Real': views from the real-robot, i.e. evaluation inputs to the policy; `NeRF': train time NeRF rendered images. \textbf{Right:} time lapse of a result on a task where the robot has to push a ball towards the target region in front of the red cones. The policy was trained in simulation by overlaying simple rendering of an orange ball on top of the scene's NeRF rendering.
           }
            \label{fig:results_fig}
        \end{center}
    }]
}

\begin{document}

\maketitle

\thispagestyle{empty}
\pagestyle{empty}

\begin{abstract}
We present a system for applying sim2real approaches to ``in the wild” scenes with realistic visuals, and to policies which rely on active perception using RGB cameras. Given a short video of a static scene collected using a generic phone, we learn the scene's contact geometry and a function for novel view synthesis using a Neural Radiance Field (NeRF). We augment the NeRF rendering of the static scene by overlaying the rendering of other dynamic objects (e.g. the robot's own body, a ball). A simulation is then created using the rendering engine in a physics simulator which computes contact dynamics from the static scene geometry (estimated from the NeRF volume density) and the dynamic objects' geometry and physical properties (assumed known). We demonstrate that we can use this simulation to learn vision-based whole body navigation and ball pushing policies for a 20 degrees of freedom humanoid robot with an actuated head-mounted RGB camera, and we successfully transfer these policies to a real robot. Project video is available at \url{https://sites.google.com/view/nerf2real/home}.
\end{abstract}

\section{Introduction}

Thanks to progress in large-scale deep reinforcement learning and scalable simulation infrastructure, training control policies in simulation and transferring them to real robots (sim2real) has become a popular paradigm in robotics~\cite{sim2real-anymal1,sim2real-anymal2,akkaya2019solving,anderson2021sim}. This approach avoids many of the issues such as state estimation, safety, and data efficiency which make it challenging to learn directly on hardware. However, creating accurate and realistic simulations is time consuming. Therefore, for sim2real to live up to its full potential, we must make it easier to recreate real scenes in simulation while accurately modelling how robots sense and interact with the world.

Reducing the gap between simulation and the real world often involves the collection of small amounts of data followed by manual tuning, the use of established system identification tools, or more recently by learning neural network models of parts of the system, e.g.~\cite{sim2real-anymal1}. It is especially difficult to accurately model the geometry and visual appearance of unstructured scenes which affect how the robot makes contact with the world and how it senses its surroundings e.g. when using a RGB camera. The need for modeling RGB cameras can partially be alleviated by using depth sensors or LiDARs which are easier to simulate and thus have a smaller sim2real gap, but such a compromise can restrict the set of tasks a robot can learn. Existing approaches to photorealistic scene reconstruction and rendering, e.g. those used for the creation of the datasets in~\cite{chang2017matterport3d, xia2018gibson, straub2019replica, szot2021habitat}, work poorly in outdoor scenes and use specialized 3D scanning setups which are not widely available, hence limiting their applicability.

In this paper we begin to address some of these challenges, and describe a system for the semi-automated generation of simulation models for visually complex scenes with highly realistic rendering of RGB camera views and accurate geometry, primarily using videos from commodity mobile cameras. To this end, we take advantage of recent advances in neural scene representations using Neural Radiance Fields (NeRF)~\cite{nerf_main, mip_nerf_360}. NeRFs are a fast developing class of scene representations that allow synthesizing novel photorealistic views from a sparse set of input views. Unlike prior work, NeRFs can be learned directly from videos or photographs from commodity mobile devices and admit access not just to a rendering function but also to the underlying scene geometry. They can be trained within minutes~\cite{mueller2022instant, kilo_nerf}, work in both indoor and outdoor settings and scale well even to large scenes such as city blocks~\cite{tancik2022block}. Together with extensions to handle dynamic scenes~\cite{pumarola2021d}, deformable objects~\cite{park2021nerfies}, and scene decompositions with novel re-combinations of objects~\cite{stelzner2021decomposing}, NeRFs can enable a general system for recreating the visuals of real-word scenes in simulation.

\begin{figure*}[t!]
\centering
  \includegraphics[width=\textwidth]{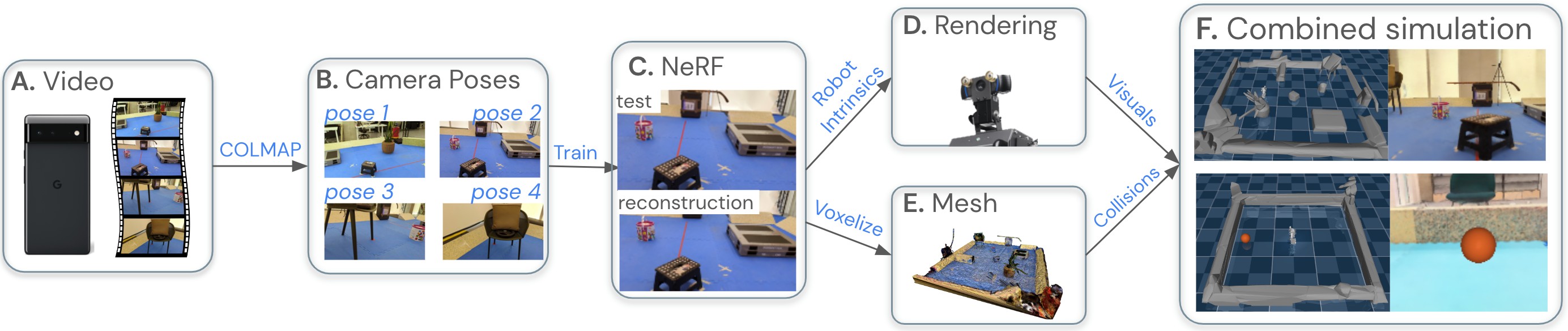}
  \caption{Overview of our system for recreating a scene in a simulator. \textbf{A.} We collect a video of the scene using a generic phone. \textbf{B.} We use structure-from-motion software to label a subset of the video with camera poses. \textbf{C.} We train a NeRF on these labeled images. \textbf{D.} We render the scene from novel views using the calibrated intrinsics of the robot's head-mounted camera. \textbf{E.} We use the same NeRF to extract the scene geometry as a mesh. We coarsen the mesh and replace the floor with a flat primitive. \textbf{F.} We combine the simplified mesh with a model of a robot, and any other dynamic objects, in a physics simulator. See Fig.~\ref{fig:object_rendering} for further details on this step.}
  \label{fig:approach}
  \vspace{-0.6cm}
\end{figure*}

Our primary contribution is an approach for combining NeRF scene representations, specifically the rendering and static geometry, learned from short (5-6) minute videos of a scene, with a physics simulation of dynamic objects such as a robot and a ball whose physical and visual properties are assumed known (see Fig.~\ref{fig:approach}). We present a semi-automated pipeline for setting up these simulations and demonstrate that they have high enough fidelity to enable simulation-to-reality transfer of vision-guided control policies. Specifically, we use a physically accurate simulation of a 20 degree-of-freedom Robotis OP3 humanoid robot together with the NeRF and end-to-end deep reinforcement learning to train vision-based whole-body navigation and ball pushing policies and we show a strong alignment between the performance of these policies in simulation and when transferred zero-shot to real robot (see Fig.~\ref{fig:results_fig} for a visualization of our results).

\section{Related work}
\subsection{Neural Radiance Fields}
Neural Radiance Fields (NeRF)~\cite{nerf_main} have recently become popular as an implicit scene representation capable of synthesizing novel photorealistic views. NeRF and its variants can represent accurate scene geometry~\cite{ref_nerf}, capture large scenes ~\cite{tancik2022block} and can be trained from photo collections without the need for localization or specialized hardware~\cite{wang2021nerf, sucar2021imap,  martin2021nerf}. Recent work has also shown training and rendering can be extremely fast and efficient~\cite{kilo_nerf, mueller2022instant}, or even be real-time on commodity handheld devices~\cite{chen2022mobilenerf}.

\textbf{NeRF in Robotics:} NeRFs have been used in robotics for pose estimation~\cite{yen2021inerf}, representation learning~\cite{yen2022nerf}, grasping~\cite{ichnowski2021dex} and dynamics model learning~\cite{driess2022learning}. There is also closely related work on obstacle avoidance within simulated NeRF environments leveraging a traditional state estimation and planning pipeline~\cite{adamkiewicz2022vision}, but transfer to the real-world was not explored. Unlike this work we use reinforcement learning to train a policy that tightly integrates perception and control for a bipedal robot and demonstrate transfer to the real world on both visual navigation and object interaction tasks.

\subsection{Visual Navigation with (Visually) Realistic Simulators}
There is a long line of work on modeling real-world indoor scenes including datasets such as Matterport3D~\cite{chang2017matterport3d}, Gibson~\cite{xia2018gibson},  Replica~\cite{straub2019replica} and Habitat-Matterport3D~\cite{ramakrishnan2021habitat}. Unlike these datasets, which were predominantly created using purpose-built scanning setups, often with access to depth or LIDAR, we use a small amount of video data from off-the-shelf mobile cameras to train a NeRF to represent our scene. 

\textbf{Visual Navigation in simulation:} Several simulation suites have been proposed for embodied visual navigation tasks combining 3D simulators with the different 3D scene datasets mentioned previously, such as Habitat~\cite{savva2019habitat, szot2021habitat}, iGibson~\cite{shen2021igibson} and AI2/ROBO-THOR~\cite{kolve2017ai2, deitke2020robothor}. These simulators have been used for learning visual navigation policies~\cite{anderson2018evaluation, wijmans2019dd, chen2019learning}, solve object-based navigation~\cite{khandelwal2022simple, batra2020objectnav}, also incorporating language commands~\cite{anderson2018vision}. These approaches primarily consider dynamically simple platforms (wheeled robots) and operate purely in simulation. 

\textbf{Sim2Real for Visual Navigation:} 
Several works have demonstrated that policies trained in these photorealistic simulators can be transferred to real-world robots. 
The majority focuses on wheeled-base robots~\cite{anderson2021sim, truong2021bi}, but some recent work has extended this to quadrupeds~\cite{truong2021learning, fu2022coupling}. These approaches use RGB-D sensors and/or LiDAR, assume access to localization, and in case of the quadrupeds, work on top of existing low-level controllers; all these assumptions help reduce the sim2real gap and thereby result in good transfer performance. In contrast, we successfully transfer whole-body vision-based control policies for a bipedal robot using only an RGB camera (which has a high sim2real gap) without access to localization or low-level controllers.

\subsection{Sim2Real in Robotics}
Sim2real transfer has made it possible to use reinforcement learning to solve several challenging real-world control problems.
Careful system identification and techniques such as Domain Randomization~\cite{tobin2017domain}, Domain Adaptation~\cite{bousmalis2017unsupervised} and Real2sim~\cite{chebotar2019closing} have helped to reduce the discrepancies between simulation and reality for the system dynamics and sensor model, enabling successes on tasks such as Rubik's cube solving with a dexterous hand~\cite{akkaya2019solving}, grasping~\cite{bousmalis2018using}, stacking~\cite{lee2021beyond}, autonomous flight~\cite{sadeghi2016cad2rl}, quadruped~\cite{hwangbo2019learning, Lee_2020, peng2020learning, sim2real-robotnpmp} and biped locomotion~\cite{yu2019sim, li2021reinforcement, siekmann2021blind}. We rely on many of the lessons learned in these works, and propose a system for high-fidelity replication of real-world scenes in simulation with which we demonstrate successful zero-shot transfer of complex vision based policies on a 20 DoF humanoid robot.

\section{Integrating NeRF with a physics simulator}
Fig.~\ref{fig:approach} presents an overview of our approach for recreating a static scene in simulation, and its extension to scenes with simple dynamic objects. Our approach consists of 6 steps: video recording, localization, NeRF training, post-processing to extract a rendering function and a collision mesh, and combining these with a physics simulator to create the simulation (see Fig~\ref{fig:object_rendering}). We describe each step below. 

\subsection{Capturing a video of the real world scene}
We capture a short $\sim5-6$ minute video of the scene using an off-the-shelf mobile camera (Google Pixel 6's rear camera in this work). A human operator walks around the scene and captures it from different viewpoints while ensuring that the camera moves slowly and evenly to reduce motion blur and minimize drastic viewpoint changes. For consistent lighting we set the white balance and brightness to a fixed (arbitrary) value. We found that high-resolution ($\geq$1080p) videos led to better localization and improved NeRF results. 

\subsection{Localization}
Next, we extract $N\sim1000$ keyframes from the captured video. We use COLMAP~\cite{schoenberger2016sfm, schoenberger2016mvs, schoenberger2016vote}, an open-source Structure-from-Motion (SfM) package, to estimate the intrinsics of the camera, and extrinsics for each keyframe (see \ref{sec:colmap_details} for details).

\subsection{NeRF training}
\label{sec:nerf_training}
Given a dataset of images and corresponding camera poses, we train a NeRF~\cite{nerf_main} to render the scene from novel viewpoints. We use recent NeRF extensions for better reconstructions, improved reconstructed geometry, and decreased rendering times. To avoid artifacts while rendering at low resolutions, we sample the average of the volume over a normal distribution~\cite{mip_nerf}. We use a space squashing formulation to support large capture areas, as well as a separate 'proposal' network, and a 'distortion' loss that encourages compact representations~\cite{mip_nerf_360}. To improve the reconstructed geometry we optimise a separate specular and diffuse color~\cite{ref_nerf}.

NeRF rendering can be compute intensive even at low resolutions. While this is not critical in our context as we use the NeRF only for offline learning in simulation, to allow for faster experiment turnaround we implement a multi scale spatial hash grid approach~\cite{mueller2022instant}. This provides a significant speedup (order of magnitude), enabling rendering one frame in 6ms on a V100 GPU. We use a similar architecture as described in ~\cite{mueller2022instant}, adding a layer normalization \cite{layernorm} before the final MLP layer, and use swish activations \cite{swish} rather than ReLU activations. Additionally, we adapted this approach to allow sampling the radiance volume over a distribution. We blur training samples with a Gaussian blur with a random variance $\sigma_{blur} \in [\sigma_{min}, \sigma_{max}]$, and provide $\Sigma = \Sigma_{sample} * (1 + (\sigma_{blur} - \sigma_{min}))$ as an extra input to the final MLP \cite{nerf_tex}. This augmentation allows the network to interpolate samples in scale-space and improves our reconstruction significantly at lower resolutions ($\sim 31.5$ vs $\sim 35.4$ average PSNR in a few held out images). NeRF hyperparameters are listed in Sec.~\ref{tbl:nerf_params}.

\subsection{Rendering in Simulation} \label{sec:rendering_in_simulation}
The trained NeRF can be used for rendering the scene from novel viewpoints and camera intrinsics. In particular, we will use it to model the robot's camera (Logitech C920) which is \textbf{different from the one used to collect images for training}. We match camera intrinsics between sim and real by calibrating the robot's camera, and use the obtained focal length and distortion parameters to render the NeRF. 

\subsection{Collision mesh extraction} \label{sec:collision_mesh_extraction}

The NeRF learns a function to predict the radiance and occupancy in space, i.e. the underlying scene geometry. We voxelize the predicted occupancy and compute a mesh via the marching cubes algorithm~\cite{lorensen1987marching}; this mesh is used for collisions within our simulation.

The camera poses obtained from COLMAP, and hence also the collision mesh vertices, are expressed in an arbitrary reference frame (including an arbitrary scale). Therefore, we need to estimate a rigid transformation and scale between this frame of reference and the simulator's world frame. We do this by solving a least-squares optimization that constrains the normal vector to the dominant floor plane in the mesh to be aligned with the z-axis in the simulator. We use Blender~\cite{blender} to manually select points on the mesh's floor for this purpose. We then manually rotate the mesh around the z-axis to a desired alignment with the simulator's world frame and compute the relative scale between the NeRF and the world by comparing the size of an object within the mesh and the real world. We also replace the floor vertices in the mesh (which can have artifacts due to a lack of texture) with a flat plane. Lastly, for faster collision computation, we crop the mesh to the extents needed for simulation. See Table~\ref{tbl:mesh_processing} for details.

\subsection{Physics simulation}

We use MuJoCo~\cite{todorov2012mujoco} as our physics simulator. The simulated scene consists of the mesh extracted from NeRF attached to the world frame as a fixed object which can collide with the robot body or other simulated dynamic objects such as a ball. Physical and visual properties of these additional virtual objects are assumed known.

\begin{figure}
  \includegraphics[width=\columnwidth]{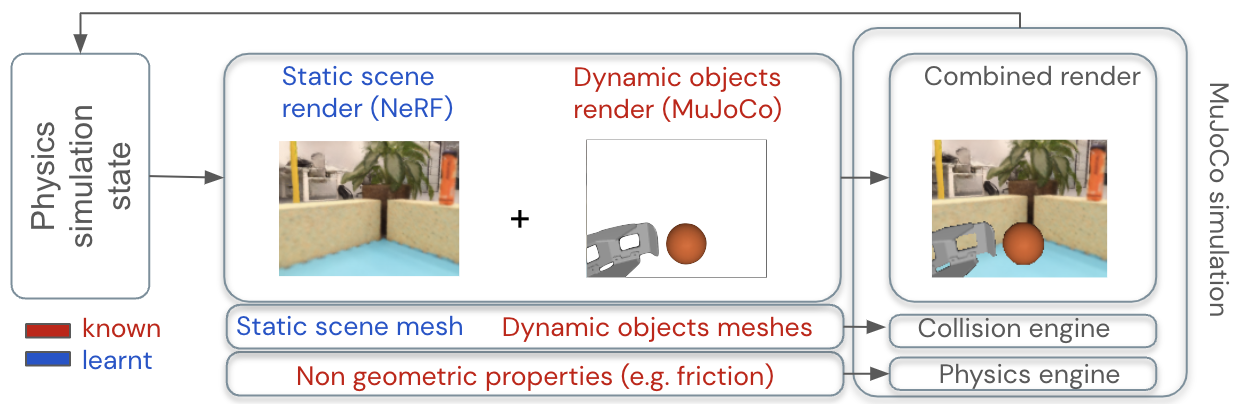}
  \caption{Our MuJoCo simulation is created by combining: (1) the learnt static scene mesh (Section \ref{sec:collision_mesh_extraction}), (2) the dynamic object meshes and (3) the learnt static scene NeRF rendering (Section \ref{sec:rendering_in_simulation}) on which (4) the Mujoco rendering of dynamic objects (a ball and robot's left arm in the camera image above) are overlaid. Other dynamic parameters (e.g. friction) are assumed known or measured.}
  \label{fig:object_rendering}
  \vspace{-0.25cm}
\end{figure}

Realistic rendering of such composite scenes is an active area of research~\cite{yang2021learning}. We opt for a straightforward approach which is suitable for our tasks. We assume that the dynamic objects (all rendered with the MuJoCo built-in renderer) are always in the foreground of the static scene (rendered with NeRF); note that this doesn't handle occlusions. Under this assumption the combined rendering is obtained by overlaying the dynamic objects rendering on top of the NeRF rendering (see Fig. \ref{fig:object_rendering} for a visualization).

\section{Sim2real training setup with NeRF + MuJoCo}
Once the combined MuJoCo simulation is set up we can train a policy purely in this simulation and deploy it directly in the real-world. We describe our training setup below.

\subsection{Humanoid Robot platform}
We use a Robotis OP3~\cite{op3} robot for all our experiments. This low cost platform is a small humanoid (about 35 \si{cm} tall, 3.5 \si{kg} in weight) with 20 actuated degrees of freedom (see Fig.~\ref{fig:results_fig}). Actuators, and hence our learned policies, are operated in a position control mode with both D and I gains set to 0. Our policies run at 40Hz and rely on the robot's on-board computer (2-core Intel NUC i3) and on-board sensors only. These include joint encoders, gyroscope, accelerometer, and a Logitech C920 camera attached to the robot's head which is actuated via two joints attached to its torso. The gyroscope and accelerometer data are filtered at 125Hz to obtain an estimate of the gravity direction in the robot's body frame using a Madgwick filter~\cite{madgwick2010efficient}. To encourage smoother movements, we apply an exponential filter with strength 0.8 to the control signals before passing them to the actuators. 

\subsection{Reducing the dynamics sim2real gap}
We ensured that the robot's sensors and actuators are modeled accurately in simulation. Specifically, we ensured that the simulated gyroscope and accelerometer data are low-pass filtered in the same way in simulation as on the real IMU chip. With these accurate models we found that policies trained in simulation transferred well to the real robot; hence we used only limited domain randomization on top of these models. Particularly, we applied random pushes to the robot during training. We also applied constant delays per episode, sampled uniformly in the range of 10\si{ms} - 50\si{ms} as well as a 5 \si{ms} jitter to all simulated sensor data to reflect various latencies on the robot. At the beginning of each episode, we attach a random mass (up to 0.5 \si{kg}) to a random position on the robot's torso and randomize the IMU's position on the torso (we shift it by up to 0.5 \si{cm}, and tilt it by up to 2 degrees). In tasks with a ball, we additionally randomize the ball's mass (0.5 - 0.9\si{kg}) and radius (11.5 - 12.5\si{cm}) at the start of each episode (the real ball weighs 0.651\si{kg} with a radius of 12\si{cm}).

\subsection{Regularizing policy learning for better sim2real transfer}
\label{sec:regularization}
Carefully choosing rewards for regularizing the robot's behavior is important for successful transfer. In all of our tasks, we use the following reward components as a regularization: 1. a constant penalty whenever the robot's yaw angular speed is larger than $\pi$ \si{rad.s^{-1}} to encourage the robot to turn slowly; 2. L2 regularization on joint angles towards a default standing pose; and, 3. a walking reward encouraging the average of feet velocities in the robot's forward direction to be 0.3 \si{m.s^{-1}}. These rewards encourage the agent to learn gaits that transfer better, and also encourage better exploration for faster learning. See Section~\ref{sec:rewards} for an exact specification of these rewards.

\subsection{Tasks}
\label{sec:tasks}
\begin{figure*}
    \centering
    \includegraphics[width=\linewidth]{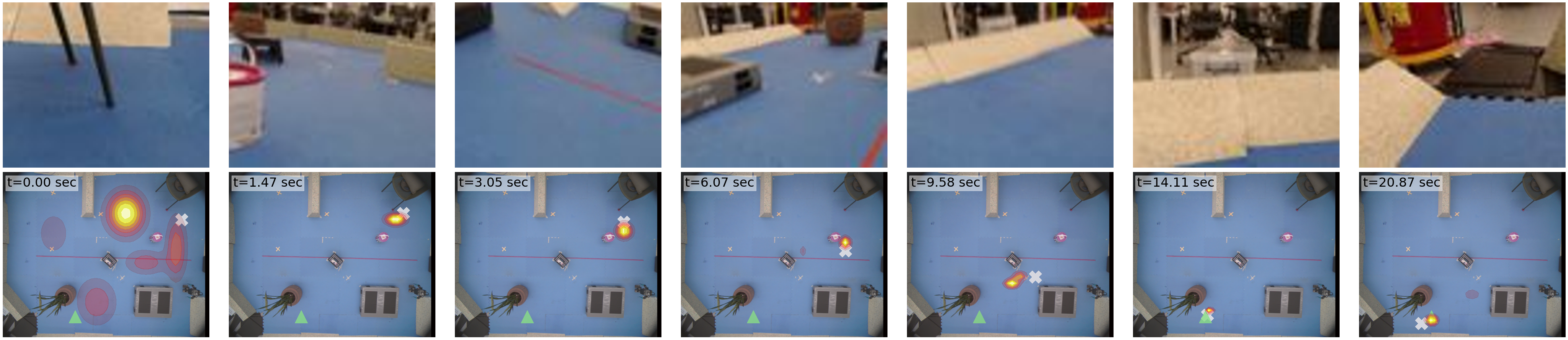}
    \caption{Localization performance of the agent. Top row: Robot camera images from a successful zero-shot transfer trial. The target is next to the potted plant. Bottom row: Visualization of the robot's belief over it's 2D position in the scene, shown as a heatmap on a top-down view of the scene. \textbf{White X:} Ground truth position from motion capture (not input to policy); \textbf{Green Triangle:} Target position.}
    \label{fig:localization_plot}
    \vspace{-0.3cm}
\end{figure*}

To demonstrate that our approach can scale to realistic scenes with complex geometries and supports simple object interactions we choose two tasks:

\textbf{Navigation and obstacle avoidance: } We demonstrate our approach on a point to point visual navigation task where the robot has to reach multiple goals (specified as (x,y) co-ordinates) 
while avoiding different obstacles such as a large plant, a chair, and walls; see Fig.~\ref{fig:results_fig} (left) and Fig.~\ref{fig:localization_plot} (bottom) for a visualization of our scene which measures 5m x 4m. 

We chose three targets in different parts of the space that the robot has to reach. We automatically compute the free areas of the scene using the NeRF's mesh and, during simulation, we randomly initialize the robot to a position and orientation within these areas. 

The \textit{reward} for training consists of the regularization terms described in Section \ref{sec:training}, and two task-specific terms: 1. a sparse bonus upon reaching the goal location; 2. a walking reward like the one we use as a regularization but instead encouraging moving in the direction of the goal at a speed of 0.3 \si{m.s^{-1}} (see Section~\ref{sec:rewards}). Episodes terminate whenever the robot's body parts other than the feet touch the scene's mesh. We consider an episode to be successful if the robot gets to $\leq$25cm of the target without falling and does not collide with any obstacles.

\textbf{Ball pushing: } As a proof of concept that we can combine static NeRF scenes with dynamic interacting objects, we consider a task in which the robot has to move a basketball to a corner of a 3m x 3m workspace (see Fig. \ref{fig:results_fig}, right). We model the basketball as a simple orange ball ignoring the fine black print which is barely visible at the resolutions we use (see Fig.~\ref{fig:object_rendering} for an example simulated image). During training in simulation, each episode starts with the ball and robot randomly positioned. In half of all episodes, we initialize the ball just in front of the robot to speed up learning. 

We again use the regularization terms in Section~\ref{sec:training} and two task-specific terms: 1. a reward for minimizing the distance between the ball and the goal region; and, 2. a reward for minimizing the distance between the robot and the ball if the ball is not moving towards the goal (see Section~\ref{sec:rewards}). Episodes are terminated whenever the robot falls. We consider an episode as successful if the robot gets the ball to the correct 1m x 1m corner square within 60 seconds. This task is much more challenging than navigation due to significant partial observability \& interactions; the robot has to search for the ball, localize itself and the ball, and move it to the goal.

\begin{figure}
  \includegraphics[width=\columnwidth]{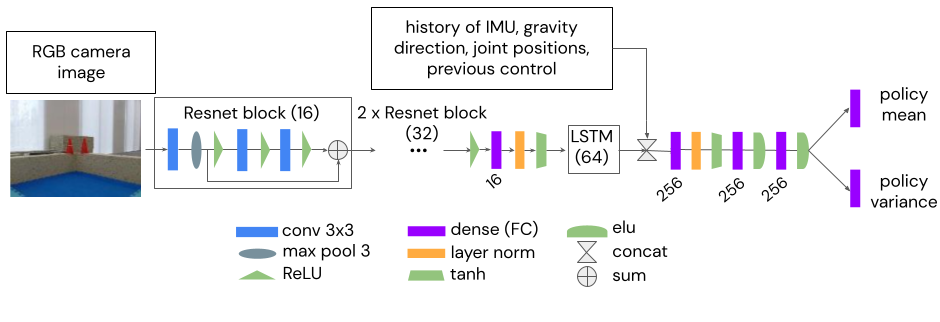}
  \caption{The policy's network architecture.}
  \label{fig:algo}
  \vspace{-0.75cm}
\end{figure}

\subsection{Policy training}
\label{sec:training}

All our policies are trained using DMPO~\cite{hoffman2020acme}, a state-of-the-art algorithm which combines distributional deep reinforcement learning~\cite{bellemare2017distributional} and MPO~\cite{abdolmaleki2018maximum, abdolmaleki2018relative}. Our policies take vision and proprioception as input; the policy network (see Fig.~\ref{fig:algo}) consists of a recurrent image encoder (to handle partial observability) which passes the RGB camera images (30x40 or 60x80 resolution) through a small ResNet followed by a LSTM. The encoded images are then combined with a history of past 5 proprioceptive observations (gyroscope, accelerometer, gravity direction estimate, joint positions, and previous control signal) and passed through an MLP which outputs a diagonal Gaussian for sampling actions. Hyperparameters used for training are listed in Section~\ref{sec:dmpo}.

We use an asymmetric actor-critic setup for training in simulation where the critic, a separate neural network that is not evaluated on the robot, receives privileged information. Specifically, the critic shares the same network structure as the actor but we replace the image encoder with the simulation's ground truth state (robot/object poses and velocities). This step is crucial for efficient learning in simulation.

\textbf{Image augmentations: } 
While the NeRF significantly reduces the sim2real gap with realistic scene renderings, we cannot easily modulate image intensity properties such as brightness or gain  (see Fig.~\ref{fig:results_fig} (left) for comparison of real vs rendered images). Thus, we apply image augmentations during training: we randomize the brightness, saturation, hue, and contrast, and apply random translations to the image, see Section~\ref{sec:img_aug} for details.

\section{Experimental results}
We now present our results. We highly encourage the reader to watch the accompanying video at \url{https://sites.google.com/view/nerf2real/home}.

\subsection{Training time}
We train policies in simulation and transfer them zero-shot to the robot for evaluation. Given $\sim$1000 frames from a 5-6 minute video, COLMAP localization takes about 3-4 hours. Training the NeRF on this dataset takes about 20 minutes on a cluster of 8 V100s, though this could be significantly sped up further~\cite{mueller2022instant}. Mesh extraction and post-processing for setting up the simulation takes a few hours, and finally training the policies takes around 24 hours (about 8M gradient updates, and 128M environment steps) for navigation and twice as long for ball pushing.

\subsection{Navigation and obstacle avoidance results}
\label{sec:navigation_results}
\textbf{Evaluation setup \& metrics:} We compare the performance of learned goal-conditioned policies in simulation to zero-shot transfer performance in real. We use the following evaluation protocol: for each of the 3 goals we chose 3 unique initial positions and 3 orientations, forming a total of 27 combinations. We perform two trials for each combination, for a total of 54 real-world episodes. We consider an episode to be successful if the robot reaches the goal ($\leq$25 cm distance) without falling. We report the \textbf{overall success percentage}, i.e. the fraction of episodes the policy was successful, and the median \textbf{time taken to reach the target}. As the evaluation space is equipped with a motion capture system we can use this to compute the ground-truth position of the robot--this ground-truth information is only used for analysis and evaluation, not as an input to the policy.

As an analysis tool, our policies are trained together with an auxiliary prediction MLP which predicts the robot's belief about its 2D position and yaw from the policy's recurrent image embedding (we do not propagate any gradients to the policy's parameters). The belief is modelled as a mixture of five Gaussians and we use it to help us disambiguate between policy failures due to confusing visual inputs and those due to other factors (e.g. the robot tripping). We use the difference between the mean of the belief distribution and the ground-truth position when quantifying the policy's localization error which is averaged over the entire episode.

\begin{table*}[ht]
\centering
\begin{tabular}{ |c||c|c|c||c|c|c| }
 \hline
  & \multicolumn{3}{c|}{Simulation results} &  \multicolumn{3}{c|}{Zero-shot transfer results} \\
 \hline
  Policy resolution & Success & Time taken & Localization error & Success & Time taken & Localization error\\
 \hline
 30 x 40 & \textbf{73$\pm$3\%} & 10.3$\pm$0.3 sec & 0.17$\pm$0.01 m & \textbf{69$\pm$6\%} & 10.3$\pm$0.4 sec & 0.24$\pm$0.01 m \\
 60 x 80 & \textbf{86$\pm$2\%} & 10.8$\pm$0.1 sec & 0.19$\pm$0.004 m & \textbf{87$\pm$5\%} & 11.2$\pm$0.4 sec & 0.27$\pm$0.04 m \\
 \hline
 
\end{tabular}
\caption{Sim \& Real performance (with standard errors) of the trained policies on the navigation and obstacle avoidance task. Time taken \& Localization error values are median statistics across all evaluation episodes.\label{tbl:navigation_results}}
\vspace{-0.5cm}
\end{table*}

\textbf{Results:} Table~\ref{tbl:navigation_results} presents the results of our evaluation. We evaluated policies with two different input resolutions, 30x40 and 60x80; due to hardware limitations on the robot (2 CPU cores, no GPU) and the need to run a policy step within 25ms, we were unable to evaluate higher resolutions.

We draw attention to a few interesting trends:
1)~\textbf{Our policies do not exhibit a significant gap between performance in simulation and on the real robot}. On the real robot, the 60x80 policy successfully reached the goal in \textbf{47/54 episodes (87$\pm$5\%)}. The lower resolution 30x40 policy performs slightly worse and was successful in 37/54 episodes (69$\pm$6\%). This is remarkably similar to the performance of these policies in simulation. Based on monitoring the policy's belief about its pose, we estimate that about half of the failures of the 40x30 policy were due to collisions with obstacles and/or the robot falling down, and the rest due to localization failures, but it is impossible to perfectly disambiguate failure modes.
2)~The policies take \textbf{similar amounts of time to reach the goal in simulation and on the robot} demonstrating that dynamical properties of the behavior such as the gait velocity also do not suffer from a sim2real gap.
3) As expected, our policies learn representations which are informative about the robot's current pose; these transfer well to the real-world. The median localization error across all control steps and trials is $\sim$0.25m. We saw that \textbf{several failures of the policy correlated with higher localization error}, specifically on a single target near the potted plant. This was particularly evident for the 40x30 policy (0.33m for failures vs 0.23m for successes). We show a visualization of the belief for a successful trajectory in Fig.~\ref{fig:localization_plot}, which demonstrates that the agent is able to quickly localize itself accurately with $\sim$2 seconds of data and rarely loses track throughout the trial. Videos of both simulated and real-world evaluations can be found on our \href{https://sites.google.com/view/nerf2real/home}{website}.

\subsection{Ball pushing results}

\textbf{Evaluation setup \& metrics:} Similar to the navigation task, we compare the \textbf{average success} of policies between simulation and real-world. We consider two different evaluation setups in real, a \textbf{Center} setting where the ball is initialized in the center cell of the workspace (Fig.~\ref{fig:results_fig}, right) and the robot is initialized in the center of any of the eight cells near the wall with 4 different orientations per cell (32 episodes total), and a \textbf{Wall} setting where the robot is initialized in the center cell facing the target corner, and the ball is initialized near the center of one of the remaining 7 cells (except the target cell) with two trials each (14 episodes total). The target is always fixed to be the corner with the two cones (see Fig.~\ref{fig:results_fig}, right), and an episode is considered successful if the robot can get the ball to the 1m x 1m target cell within 60 seconds without falling down. 

\begin{table}
\centering
\begin{tabular}{ |c||c|c||c|c| }
 \hline
   & \multicolumn{2}{c|}{Sim Success} &  \multicolumn{2}{c|}{Real Success} \\
 \hline
   Policy resolution & Center & Wall & Center & Wall\\
 \hline
 30 x 40 & \textbf{99\%} & 100\% & \textbf{78$\pm$7\%} & 43$\pm$14\% \\
 \hline
\end{tabular}
 \caption{Sim \& real performance (with standard errors) of trained policies on the ball pushing task with ball initialized in the center of the arena vs in different cells near the wall. \label{tbl:pushing_results}} 
 \vspace{-0.75cm}
\end{table}

\textbf{Results:} Table~\ref{tbl:pushing_results} presents the evaluation results. We highlight two key points: 1) As can be seen in the accompanying video, our policies use the robot's hands to move the ball, and exhibits active perception when searching and tracking the ball. These behaviors emerge from the task requirements and are \textbf{not explicitly encouraged by the reward function}; they also transfer successfully from simulation to the real world. 2) While our policies show good performance, the sim2real gap is larger for this contact-rich task. This is especially true for the \textbf{Wall} initializations; unless the robot executes a perfect push, the ball often gets stuck near the walls and the robot has a hard time moving it. 

Videos of all our results, and comparisons of renderings from the NeRF, COLMAP reconstructions (which we show as a baseline) and real images can also be found in the supplementary material and on our \href{https://sites.google.com/view/nerf2real/home}{website}.

\section{Discussion and Limitations}
We have presented a pipeline for creating simulation environments of visually complex scenes in a way that allows training vision guided policies for sim2real transfer. To this end we combine the scene geometry and rendering function derived from a NeRF with a known physics model of the robot and (optionally) additional objects.

In principle, our approach is embodiment independent and it can be automated further in future work. For instance, new NeRF-like models such as \cite{ref_nerf, nerf_impl, nerf_neus}, may improve scene geometry reconstruction thus eliminating the need for manual postprocessing of minimally textured areas like the floor. Evaluating the approach on contact-rich tasks such as climbing on objects will allow us to better assess the current limitations of the approach, and may guide future NeRF and scene modeling developments.

We have currently opted for a very simple approach to composing the rendering of an a priori known object with the rendering of a static NeRF scene. However, the field has been actively working on NeRF-like approaches to photorealistic rendering of composite scenes \cite{guo2020object}, including ways for segmenting a static scene into dynamic objects and representing each using a separate NeRF \cite{stelzner2021decomposing}. If necessary, our pipeline can leverage any of these improvements (and will have to for more complex dynamic scenes).

Similarly, we believe that recent work on eliminating the computationally expensive 
localization step from NeRF pipelines \cite{wang2021nerf, sucar2021imap,  martin2021nerf}, and speeding up both NeRF~\cite{mueller2022instant} and RL training~\cite{rudin2021learning} will soon enable going from a video of a scene to a trained policy within minutes or hours instead of 1-2 days, potentially enabling running our setup online on the robot during deployment.

\textbf{Impact statement:}
This work presents an approach to train vision guided policies for general robotics systems. While in its current form the approach is unlikely to enable real world applications, future research may make possible a range of applications that can benefit humanity. We strongly oppose any applications designed to bring harm to humans.

\addtolength{\textheight}{0cm}   




\section*{Acknowledgments}
We would like to thank Neil Sreendra, Marlon Gwira, Kushal Patel, Nathan Batchelor, and Federico Casarini for maintaining and repairing robots used in this project, Jon Scholz and Francesco Romano for reviewing the paper, and Claudio Fantacci for helping with paper figures.

\section*{APPENDIX}
\subsection{COLMAP details}
\label{sec:colmap_details}

To train the NeRF model, we need a paired set of images and camera poses. We divide the video into $N\sim1000$ equal partitions, and for each partition we use a heuristic to pick the least motion blurred frame. We use the average variance of the Laplacian of each frame to approximate how sharp a frame is.

After extracting these keyframes, we feed them into COLMAP~\cite{schoenberger2016sfm, schoenberger2016mvs, schoenberger2016vote}. We use the sequential matcher to generate camera poses with mostly default settings, except for using affine SIFT features, guided matching, and forcing a single OPENCV style camera.

For the comparisons between the NeRF reconstruction and COLMAP reconstruction shown in the accompanying video and on our \href{https://sites.google.com/view/nerf2real/home}{website}, we reconstruct a dense mesh from the sparse results using the Poisson mesher with the default settings.

\subsection{NeRF implementation details}
\label{sec:nerf_hypers}

As described in Section~\ref{sec:nerf_training}, we use a NeRF with a similar architecture to the one used in \cite{mueller2022instant}. We use 'swish' \cite{swish} rather than 'relu' activations however, and add a final layernorm \cite{layernorm} before the last MLP layer.

\begin{figure*}[h!]
    \centering
    \includegraphics[width=\linewidth]{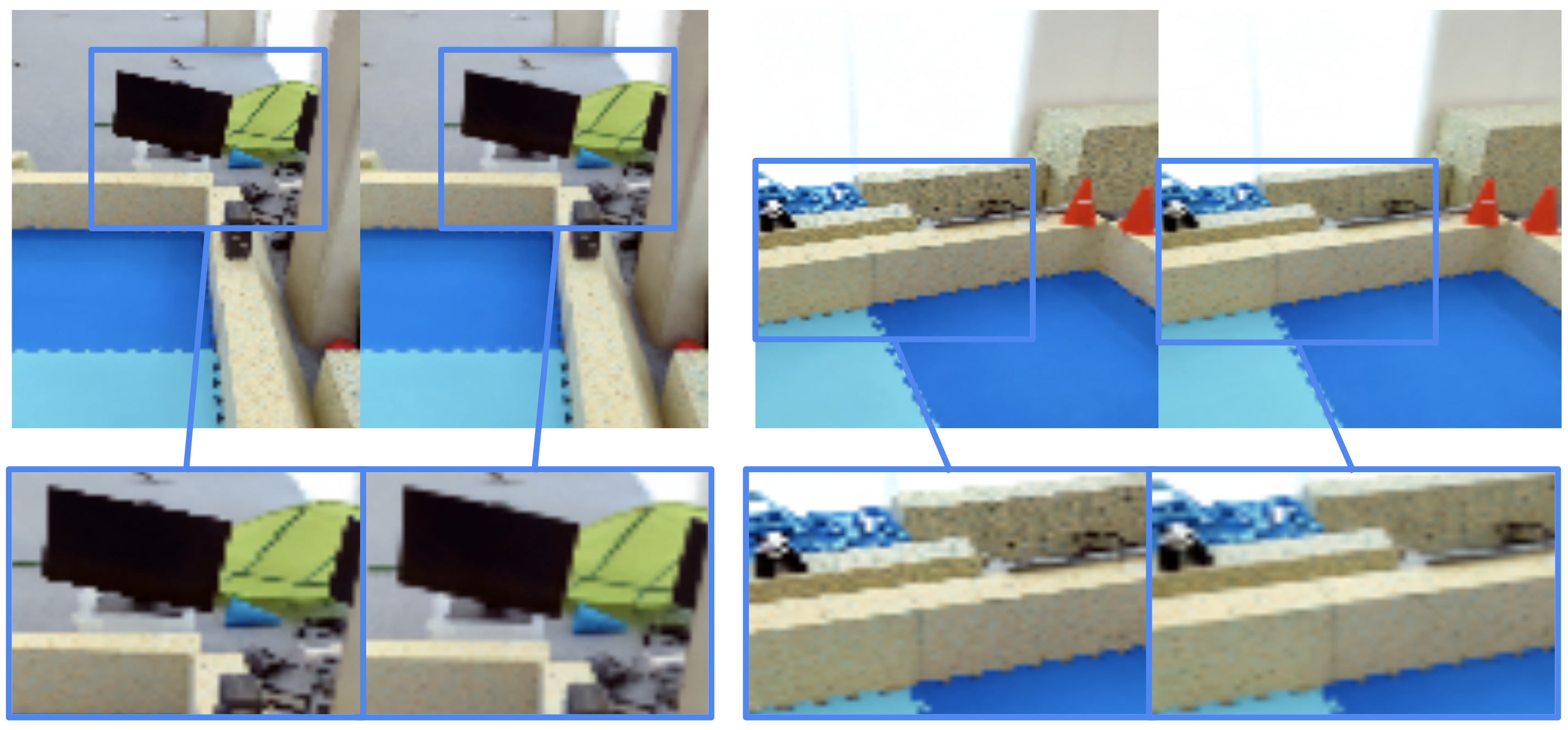}
    \caption{Two different views of one of the captured scenes. For each view we show an example rendered from a model trained without (left) and with (right) our gaussian blur augmentation. We concatenate the diagonalised variance and train with augemented samples. This allows the network to effectively interpolate in scale space. Zooming in shows the reduced 'stair-stepping' artifacts in the renders.}
    \label{fig:nerf_blur_ablation}
    \vspace{-0.2cm}
\end{figure*}

As described in Section~\ref{sec:nerf_training}, we've also integrated mip-NeRF style sampling with the hashtable style NeRF by appending the diagonalised variance to the sample position, and feeding in blurred training samples. Figure \ref{fig:nerf_blur_ablation} compares results with and without this method when rendering at lower resolutions.

Table \ref{tbl:nerf_params} lists the hyperparameters of the NeRF architecture and hyperparameters for training the model.

\begin{table}[h!]
\centering
\begin{tabular}{|l | c|} 
 \hline
    NeRF model parameters & \\
  \hline
    hashtable size      & $2^{20}$    \\
    hashtable levels    & $12$ \\
    samples             & $64 \times 64$ \\
    MLP depth           & $2$       \\
    MLP width           & $64$      \\
    MLP width viewdir   & $32$      \\
    Proposal MLP depth  & $2$       \\
    Proposal MLP width  & $64$      \\
    sample dilation     & $0.001$   \\
    activation type     & $\text{swish}$   \\
    density activation  & $\text{squareplus}$ \cite{squareplus} \\
    density bias        & $-5$      \\
  \hline
    NeRF training parameters & \\
  \hline
    batch size & 1$6384$ \\
    weight decay & $\num{5e-5}$ \\
    optimizer & $\text{adamw}$ \\
    lr init & $\num{2e-3}$ \\
    lr final & $\num{4e-5}$ \\
    warmup steps & $2048$ \\
    max steps & $40000$ \\
    charb padding & $0.001$ \\
    distortion loss mult & $0.01$ \\
    blur $\sigma_{max}$ & $12.0$ \\
 \hline
\end{tabular} \vspace{2mm}
\caption{List of NeRF hyperparameters.}
\label{tbl:nerf_params}
\end{table}

\subsection{Mesh processing details}
Table \ref{tbl:mesh_processing} visualizes the different stages of the mesh processing needed to go from the raw mesh extracted from NeRF's occupancy network to a simplified mesh suitable for MuJoCo simulation.

\begin{table*}[h!]
\begin{tabular}{| c | c| c|} 
\hline
\includegraphics[width=0.3\textwidth]{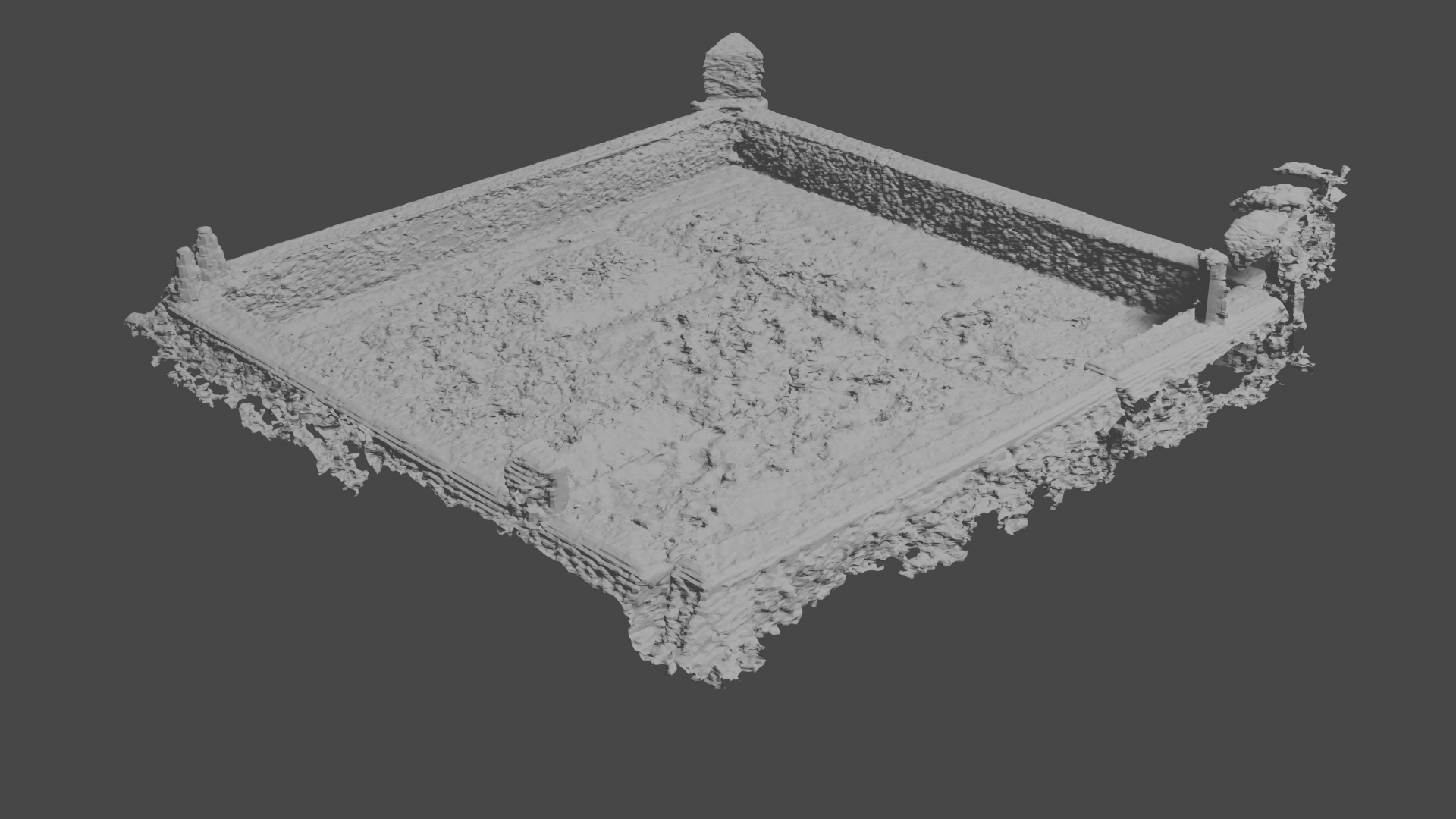} &
\includegraphics[width=0.3\textwidth]{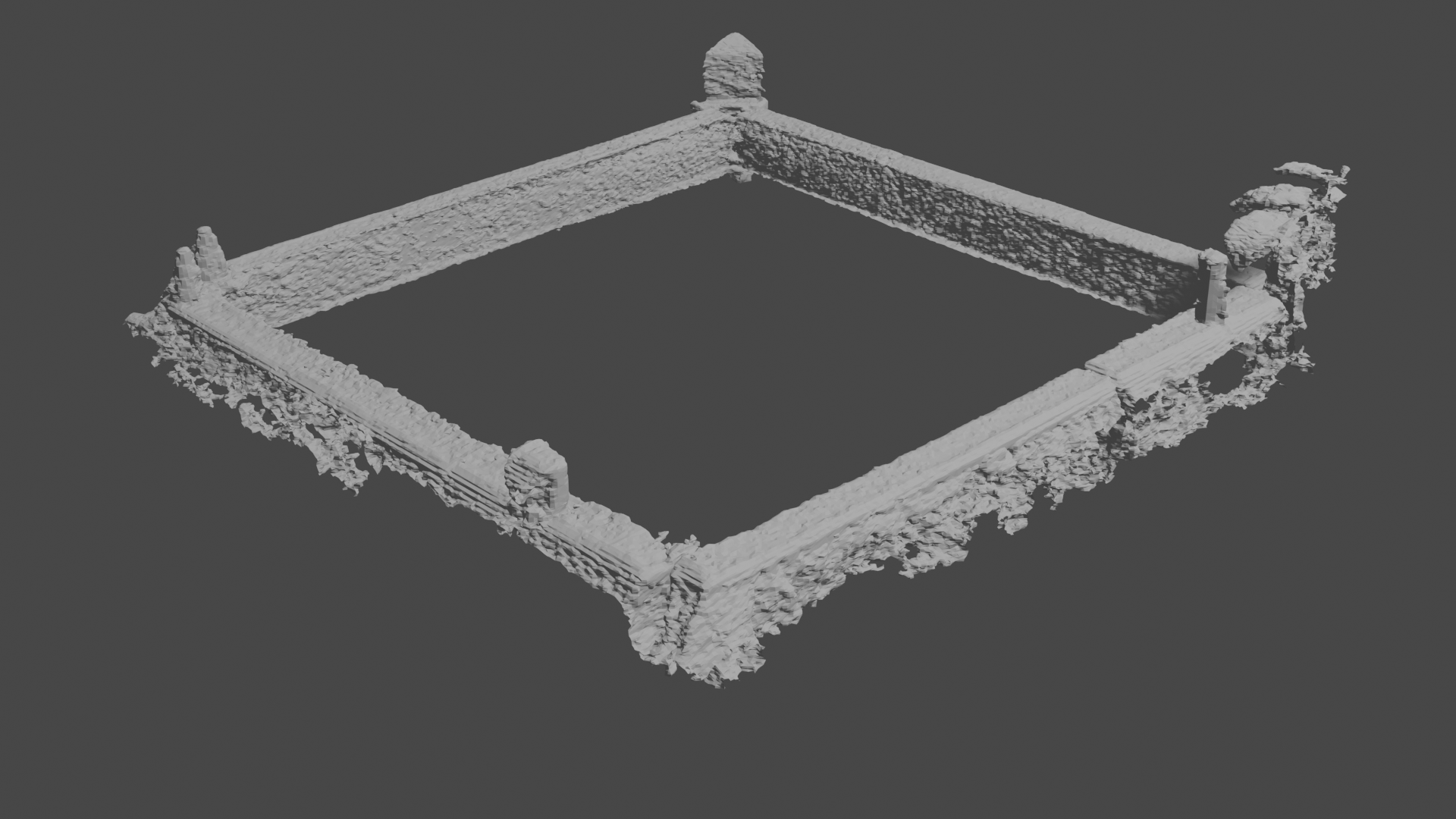} &
\includegraphics[width=0.3\textwidth]{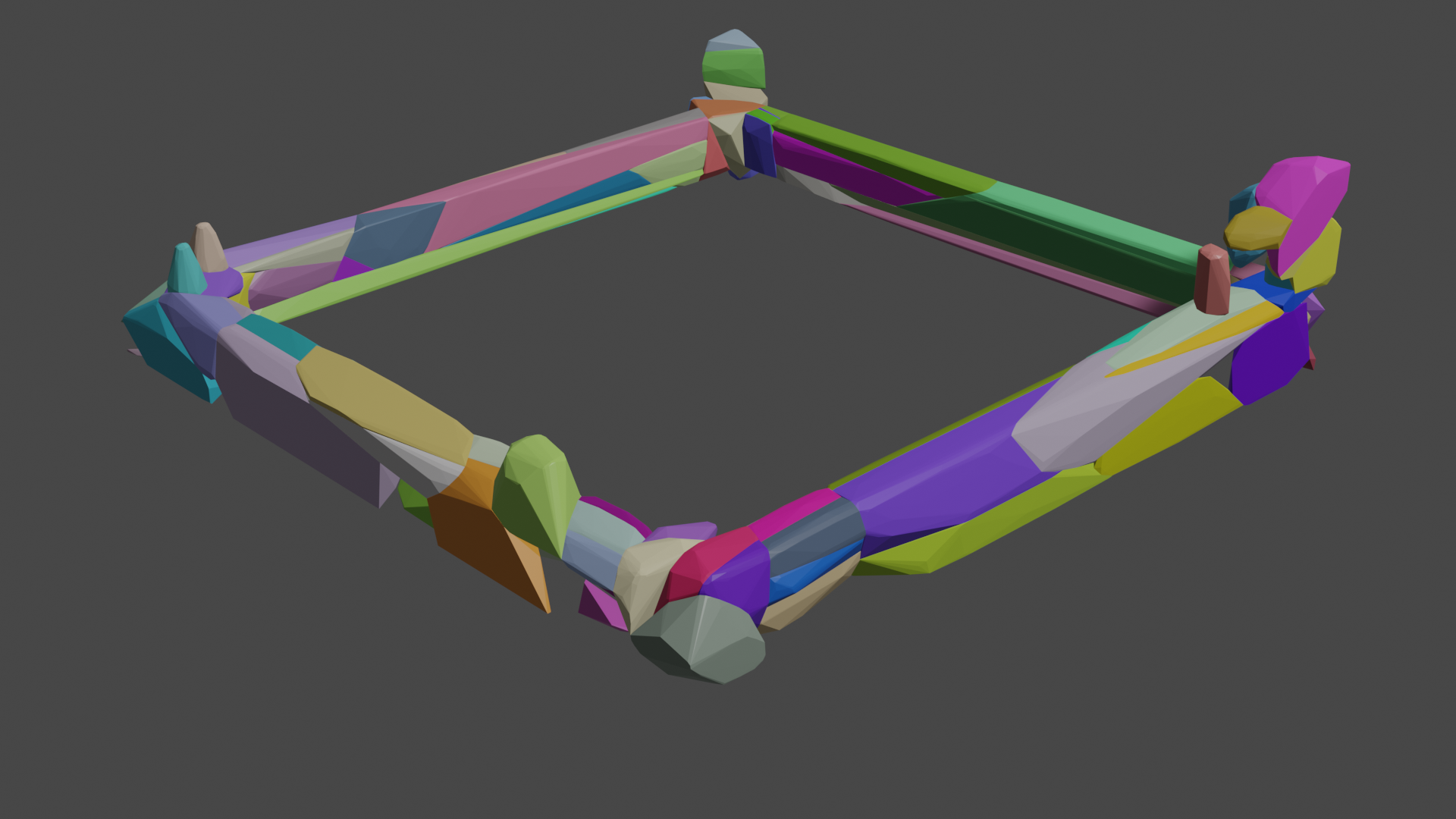}\\
\hline
\includegraphics[width=0.3\textwidth]{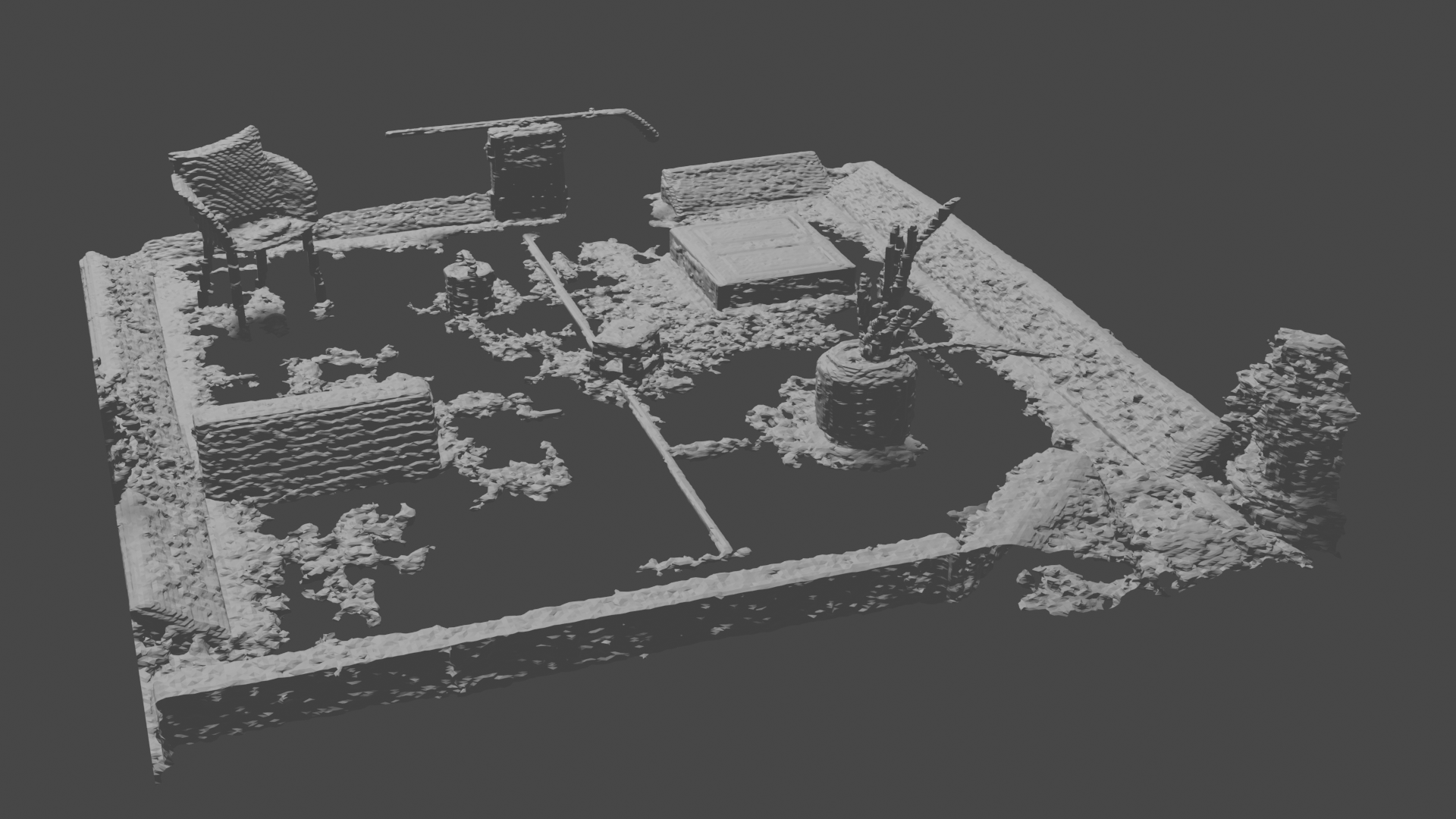} &
\includegraphics[width=0.3\textwidth]{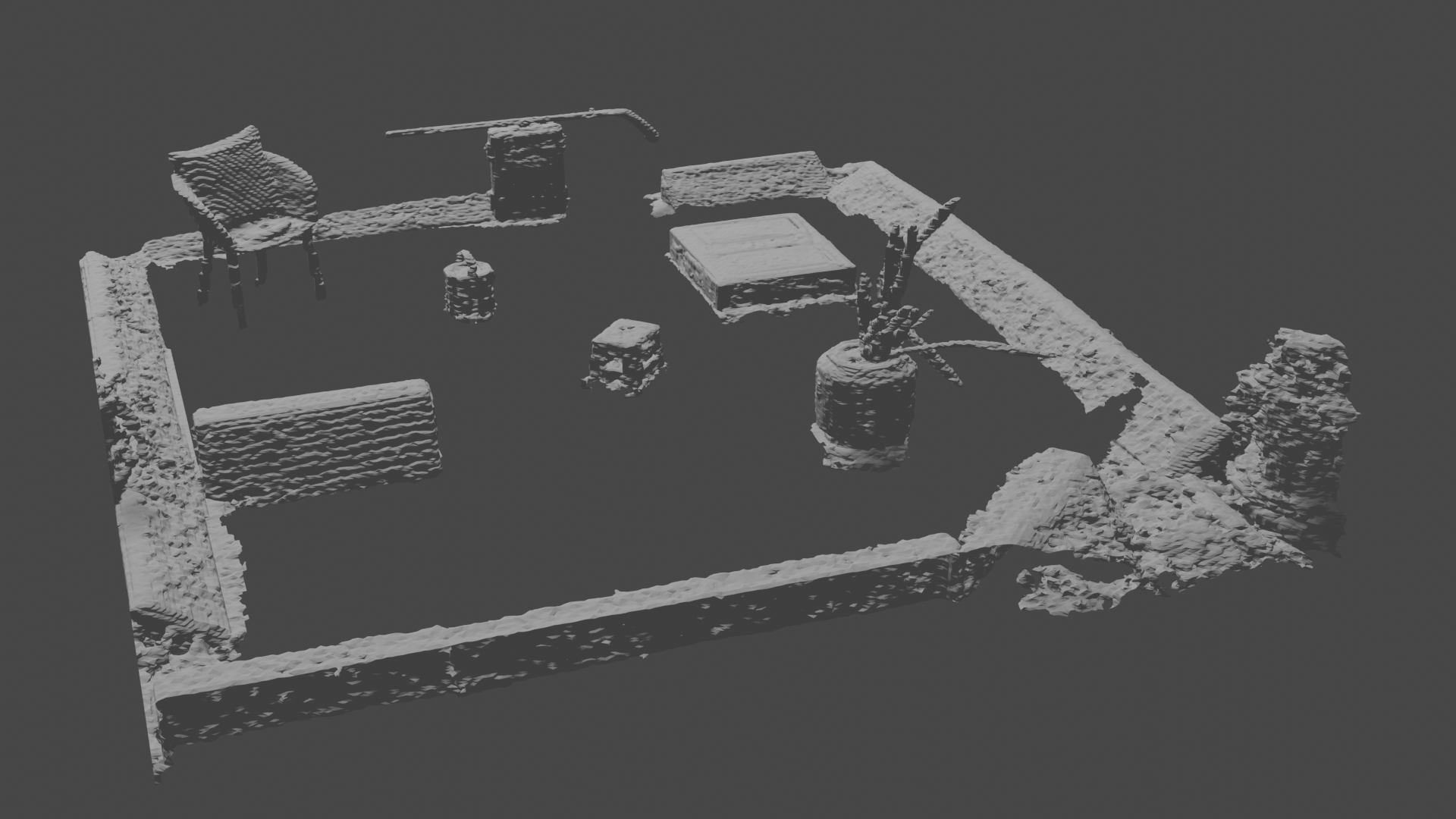} &
\includegraphics[width=0.3\textwidth]{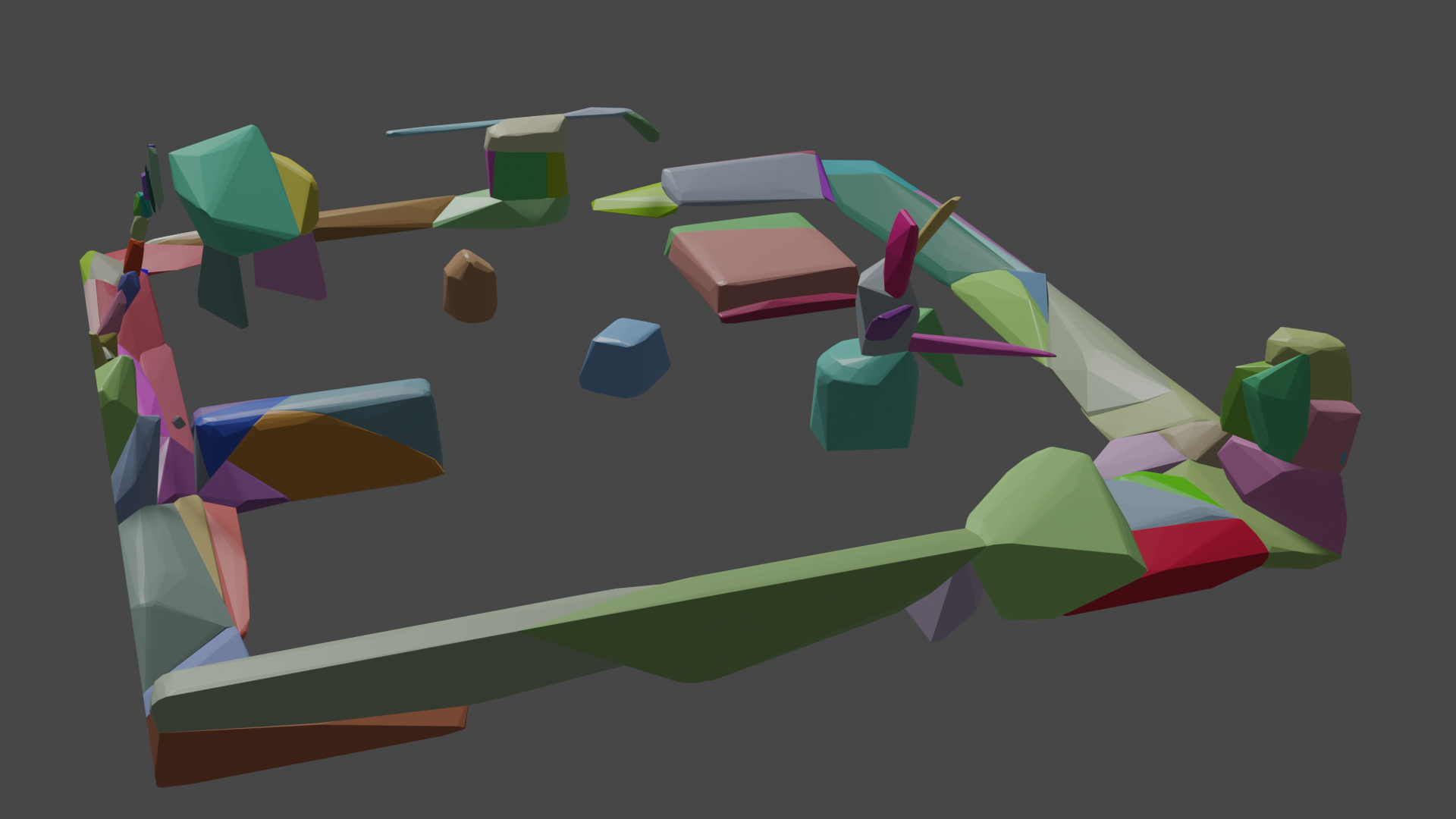} \\
\hline
\end{tabular}
\caption{\textbf{Mesh preprocessing.} Two examples of how the NeRF meshes are processed to make them suitable for simulation. The leftmost column shows the meshes of two different static environments, obtained by running marching cubes on a discretized occupancy grid from the trained NeRF and rotating, translating and cropping the resulting mesh to the extents of the scene. The central column shows the same meshes with the floor removed (as described in Section \ref{sec:collision_mesh_extraction}). Note that for the raw mesh in the bottom left, a partial floor removal was done by choosing a high threshold for the occupancy within the marching cubes procedure. In the rightmost column the mesh is decomposed into convex sub-components (each sub-components in a different colour), which are passed to MuJoCo for collision detection. The convex decomposition step is specific to MuJoCo as it does not handle collisions between non-convex objects.}
\label{tbl:mesh_processing}
\vspace{-0.25cm}
\end{table*}

\subsection{Policy training details}

\subsubsection{Control limits}

We clipped the inputs to actuator position controllers to be within the limits specified in Table~\ref{tbl:robot_pose}. For the ball pushing task, we further extended the limits of the \texttt{head\_pan} joint to be within (-2.5, 2.5) so that the policy can learn to actively look at the ball. In both tasks we found it important to limit the upper bound of the \texttt{head\_tilt} joint to prevent the policy from looking at the ceiling which is not well captured by the NeRF due to a lack of data in the input video.

\begin{table}[h!]
\centering
\begin{tabular}{|l | c| c|} 
 \hline
 \textbf{joint} & \textbf{reference pose [rad]} & \textbf{range [rad]} \\
 \hline
 head pan & $0.043$ & (-0.79, 0.79)\\ 
 head tilt & $-0.47$ & (-0.79, -0.157)\\ 
 left ankle pitch & $0.638$ & (-0.25, 1.109)\\ 
 left ankle roll & $-0.0414$ & (-0.4, 0.4)\\ 
 left elbow & $-0.7915$ & (-0.8, 0.2)\\ 
 left hip pitch & $-0.5553$ & (-0.45, 1.57)\\ 
 left hip roll & $-0.0383$ & (-0.4, -0.1)\\ 
 left hip yaw & $0$ & (-0.3, 0.3)\\ 
 left knee & $1.0646$ & (-0.2, 2)\\ 
 left shoulder pitch & $-0.0874$ & (-0.785, 0.785)\\ 
 left shoulder roll & $0.7915$ & (0, 0.8)\\
 right ankle pitch & $-0.638$ & (-1.109, 0.25)\\ 
 right ankle roll & $0.0414$ & (-0.4, 0.4)\\ 
 right elbow & $0.7915$ & (-0.2, 0.8)\\ 
 right hip pitch & $0.5553$ & (-1.57, 0.45)\\ 
 right hip roll & $0.0383$ & (0.1, 0.4)\\ 
 right hip yaw & $0$ & (-0.3, 0.3)\\ 
 right knee & $-1.0646$ & (-2, 0.2)\\ 
 right shoulder pitch & $0.0874$ & (-0.785, 0.785)\\ 
 right shoulder roll & $-0.7915$ & (-0.8, 0)\\
 \hline
\end{tabular} \vspace{2mm}
\caption{Joint reference poses and limits.}
\label{tbl:robot_pose}
\end{table}

\subsubsection{Rewards}
\label{sec:rewards}

We used the following regularization rewards (from Section~\ref{sec:regularization}) to shape the robot's behavior and improve sim2real transfer:
\begin{equation*}
    r_{\text{turn}} = -1.\ \text{if}\ \omega_{yaw} > \pi\ \text{else}\ 0.,
\end{equation*}
\begin{equation*}
    r_{\text{pose}} = \sqrt{\frac{1}{20}\sum_{i=1}^{20} \frac{(\mathbf{q}_i - \mathbf{ref}_i)^2}{\mathbf{range}_i^2}},
\end{equation*}
\begin{equation*}
    r_{\text{speed}} = \frac{0.3 - |\mathbf{v}_{x}^{\text{feet}} - 0.3|}{0.3},
\end{equation*}
where $\omega_{yaw}$ is the angular velocity of the robot's gravity-aligned body frame. The gravity-aligned body frame is obtained from the robot's torso frame by rotating it using the smallest rotation which aligns robot's frame negative z-axis with gravity direction. $\mathbf{q}$ are the robot's joint positions, $\mathbf{ref}$ are the corresponding reference poses, and $\mathbf{range}$ are the joint ranges (see Table~\ref{tbl:robot_pose} for specific values). $\mathbf{v}^{\text{feet}}$ is the velocity of the midpoint of the robot's feet expressed in the gravity-aligned body frame.

The task-specific reward components we used for the navigation task (Section~\ref{sec:tasks}) are
\begin{equation*}
    r_{\text{navigate sparse}} = 1.\ \text{if}\ ||\mathbf{x}_{\text{goal}}|| < 0.25\ \text{else}\ 0.,
\end{equation*}
\begin{equation*}
    r_{\text{navigate}} = \left(0.3 - \left\lVert\mathbf{v}_{x, y}^{\text{feet}} - 0.3\frac{\mathbf{x}_{\text{goal}}}{||\mathbf{x}_{\text{goal}}||}\right\rVert\right) / 0.3,
\end{equation*}

where $\mathbf{x}_{\text{goal}}$ is the 2d goal position in the robot's gravity-aligned frame. The full reward we used for training the navigation policies is 
\begin{multline*}
    r_{\text{navigation task}} = r_{\text{turn}} + 0.5r_{\text{speed}} + 0.5r_{\text{pose}} + \\ r_{\text{navigate sparse}} + 0.25r_{\text{navigate}}.
\end{multline*}

The task-specific reward for the ball pushing task (Section~\ref{sec:tasks}) works in two stages. If the ball is near the goal or is moving towards the goal, then we do not encourage any robot behavior. If this is not the case, then we encourage the robot to move towards the ball if it is away from it, otherwise we encourage it to stay close to the ball. Specifically, let $d_{ball} = ||\mathbf{x}_{\text{robot}} - \mathbf{x}_{\text{ball}}||$, $d_{goal} = ||\mathbf{x}_{\text{ball}} - \mathbf{x}_{\text{goal}}||$, and $v_{ball}$, $v_{goal}$ the time derivatives of these distances. Let 
\begin{equation*}
    \rho(d, v) = e^{-d^2-v^2} + \left(1-e^{-d^2}\right)\left(1-e^{-\min(0, v)^2}\right).
\end{equation*}
The task specific reward is
\begin{equation*}
    r_{\text{ball}} = \rho(d_{goal}, v_{goal}) + \left(1 - \rho(d_{goal}, v_{goal})\right)\rho(d_{ball}, v_{ball}) / 2,
\end{equation*}
and the full reward for training ball pushing policies is
\begin{equation*}
    r_{\text{ball pushing task}} = r_{\text{turn}} + 0.5r_{\text{speed}} + 0.5r_{\text{pose}} + r_{\text{ball}}.
\end{equation*}

\subsubsection{DMPO}
\label{sec:dmpo}

Our DMPO implementation is based on the open-source reference implementation \url{https://github.com/deepmind/acme/tree/master/acme/agents/tf/dmpo}~\cite{hoffman2020acme}. Readers can find more details about this algorithm in the MPO~\cite{abdolmaleki2018maximum, abdolmaleki2018relative} and distributional RL~\cite{bellemare2017distributional} papers, as well as in~\cite{bloesch2021towards} which used the exact same implementation we are using. Here we only mention the differences between ours and the linked reference implementation:
\begin{itemize}
    \item[1.] We use JAX~\cite{jax2018github} instead of TensorFlow.
    \item[2.] We use a distributed asynchronous setting with separate actor and learner processes.
    \item[3.] In order to support recurrent architectures, we calculate losses and gradients over batches of multi-step trajectories instead of batches of single-step transitions.
    \item[4.] N-step returns are estimated from multi-step trajectories inside the critic loss rather than being accumulated in the actor process.
\end{itemize}

Hyperparameters are listed in Table~\ref{tbl:dmpo}.

\begin{table}[h!]
\centering
\begin{tabular}{|l | c|} 
 \hline
 policy learning rate & $0.0001$ \\ 
 critic learning rate & $0.0001$ \\ 
 dual variables learning rate & $0.01$ \\ 
 trajectory length & $48$ \\ 
 batch size & $32$ \\ 
 updates per environment step & $1/8$ \\ 
 discount & $0.99$ \\ 
 init log temperature & $10$ \\ 
 init log alpha mean & $10$ \\ 
 init log alpha stddev & $1000$ \\ 
 epsilon & $0.1$ \\ 
 epsilon mean & $0.0025$ \\ 
 epsilon stddev & $1\text{e}-6$ \\ 
 epsilon action penalty & $0.001$ \\ 
 per dim KL constraints & True \\
 out-of-bounds action penalization & True \\
 n step & $5$ \\ 
 target actor update period & $25$ \\ 
 target critic update period & $100$ \\ 
 vmin & $-150$ \\ 
 vmax & $150$ \\  [1ex]
 \hline
\end{tabular} \vspace{2mm}
\caption{List of DMPO hyperparameters.}
\label{tbl:dmpo}
\end{table}

\subsection{Camera calibration and image augmentations}
\label{sec:img_aug}

\begin{table}[h!]
\centering
\begin{tabular}{|l | c|} 
 \hline
 \textbf{dm\_pix function} & \textbf{function arguments} \\
 \hline
 \texttt{random\_brightness} & \texttt{max\_delta=32. / 255.} \\ 
 \texttt{random\_hue} & \texttt{max\_delta=1. / 24.} \\ 
 \texttt{random\_contrast} & \texttt{lower=0.5, upper=1.5} \\ 
 \texttt{random\_saturation} & \texttt{lower=0.5, upper=1.5} \\ 
 \hline
\end{tabular} \vspace{2mm}
\caption{Image augmentation settings.}
\label{tbl:img_aug}
\end{table}

We calibrated the robot's Logitech C920 camera's focal length and distortion parameters using the \texttt{camera\_calibration} ROS package. In order to address the mismatch between the image intensity settings of the robot's camera and the camera used for data collection, we use image augmentations during policy training.
Specifically we use the \texttt{random\_brightness}, \texttt{random\_hue}, \texttt{random\_contrast}, and \texttt{random\_saturation} functions from the \texttt{dm\_pix} library~\cite{deepmind2020jax} with the parameters listed in Table~\ref{tbl:img_aug}. We also apply random translations by up to $5\%$ of the image height/width. However, these augmentations are not a replacement for calibration. We found the gain parameter of the camera to be particularly important and so we manually tuned it by comparing to the images used for training NeRF (we used gain=50 for the navigation task, and gain=128 for the ball pushing task). 

\begin{figure*}[h!]
    \centering
    \includegraphics[width=\linewidth]{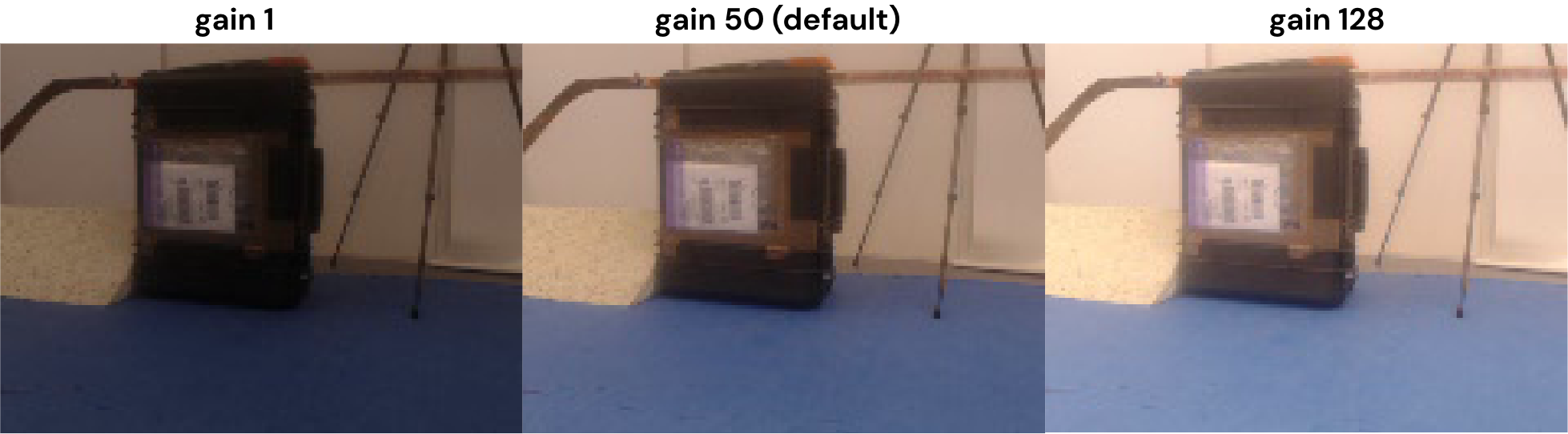}
    \caption{The effect of changing the camera gain. The policy breaks when the gain is 128, and sometimes gets lost when it is 1.}
    \label{fig:gain_ablation}
    \vspace{-0.3cm}
\end{figure*}

The sensitivity to gain is quantified in Table~\ref{tbl:gain_ablation} and the effects of varying gain on the camera images is visualized in Fig.~\ref{fig:gain_ablation}. The policy is especially sensitive to high gain values (leading to brighter images) which cause it to walk in a small circle. It is less sensitive to small gain values for which it still navigates to the target but sometimes gets lost or hits obstacles.

\begin{table}[h!]
\centering
\begin{tabular}{|l | c| l|} 
 \hline
 \textbf{Gain} & \textbf{Success rate} & \textbf{Behavior} \\
 \hline
 128 & 0/5 & Robot walks in a small circle.\\ 
 90 & 0/5 & Robot walks in a small circle.\\ 
 50 & 4/5 & Robot consistently reaches the target.\\
 && This is the default setting for our evaluation \\ 
 && results shown in Section~\ref{sec:navigation_results}\\
 10 & 1/5 & Robot often reaches the target (4/5) but\\
 && hits obstacles on the way.\\
 1 & 2/5 & Robot occasionally reaches the target but \\
 && often gets lost.\\
 \hline
\end{tabular} \vspace{2mm}
\caption{Performance of the 80x60 resolution navigation policy when varying the camera's gain parameter. Both the goal position and the robot's initial position were fixed in all trials. Higher gain corresponds to brighter image, and 50 is the default value used in the navigation experiments.}
\label{tbl:gain_ablation}
\end{table}

\subsection{Ablations and other results}

We qualitatively evaluated our navigation policies in perturbed settings such as moving some of the fixed objects in the scene or having humans to walk around the scene. These evaluation runs are shown in the accompanying videos on our \href{https://sites.google.com/view/nerf2real/home}{website}. We were surprised that the ``low-level" locomotion of the robot was disentangled from the ``high-level" scene understanding even though the policies were trained end-to-end. For example, when we changed the scene setup, the robot might walk in the wrong direction but it would retain a stable gait. Similarly, when we blocked the robot's camera, the robot would stop walking but it would not fall (and it would start walking again once we unblocked its view). Additionally, we often saw that the robot would reach the successful target even in these perturbed settings.

\bibliography{citations} 
\bibliographystyle{ieeetr}

\clearpage


\end{document}